\documentclass[11pt]{article}

\usepackage[]{acl}

\usepackage{times}
\usepackage{latexsym}

\usepackage[T1]{fontenc}
\usepackage[utf8]{inputenc}

\usepackage{microtype}

\usepackage{inconsolata}

\usepackage{graphicx}

\usepackage{booktabs, colortbl} 
\usepackage{multirow,makecell} 
\usepackage{multicol}

\usepackage{amsmath}
\usepackage{amssymb}
\usepackage{caption}
\usepackage{subcaption}
\usepackage{titlesec}
\usepackage{pgfplots}
\usepackage[normalem]{ulem} 
\usepackage{enumitem}
\usepackage{graphicx} 
\usepackage{hyperref}
\usepackage{svg}

\usepackage{enumitem}
\usepackage[splitrule]{footmisc}
\setlist[itemize]{noitemsep, nolistsep}
\usepackage{color, colortbl}
\usetikzlibrary{positioning, arrows.meta, calc,backgrounds}

\usepackage{tcolorbox}
\usepackage{listings}
\usepackage{xcolor}
\newtcolorbox{promptbox}{
    colback=gray!6,
    colframe=gray!60!black,
    fontupper=\sffamily,
    left=5mm, right=5mm,     % Innerer Abstand links/rechts
    top=4mm, bottom=0mm,     % Innerer Abstand oben/unten
    arc=4.5mm, % Abgerundete Ecken
    boxrule=0.8pt
}

\lstdefinestyle{promptstyle}{
    basicstyle=\sffamily\small, % Serifenlose Schrift für den Prompt-Inhalt
    breaklines=true,            % Automatische Zeilenumbrüche aktivieren
    columns=fullflexible,       % WICHTIG: Verhindert die feste "Code-Spaltenbreite" -> Text fließt normal
    keepspaces=true,            % Behält Leerzeichen bei
    aboveskip=4pt,              % Minimaler Abstand nach dem Box-Titel
    belowskip=0pt
}

\lstdefinestyle{codestyle}{
    backgroundcolor=\color{backcolour},   % Leicht grauer Hintergrund für Code
    commentstyle=\color{codegreen},
    keywordstyle=\color{magenta},
    numberstyle=\tiny\color{codegray},
    stringstyle=\color{codepurple},
    basicstyle=\ttfamily\small,          % WICHTIG: Monospace-Schriftart (Schreibmaschine)
    breakatwhitespace=false,         
    breaklines=true,                 
    captionpos=b,                    
    keepspaces=true,                 
    numbersep=5pt,                  
    showspaces=false,                
    showstringspaces=false,
    showtabs=false,                  
    tabsize=4,
    columns=fixed,                        % WICHTIG: Feste Spaltenbreite für Einrückungen
}
\definecolor{codegreen}{rgb}{0,0.6,0}
\definecolor{codegray}{rgb}{0.5,0.5,0.5}
\definecolor{codepurple}{rgb}{0.58,0,0.8}
\definecolor{backcolour}{rgb}{0.95,0.95,0.95}
\title{Predicting Steering Vectors and Adapter Weights for Few-Shot Author-Style Transfer}

\author{Leonard Popp \phantom{\and} Danni Liu \phantom{\and}  Supriti Sinhamahapatra \phantom{\and} Jan Niehues \\
        Karlsruhe Institute of Technology, Germany \\
        \texttt{leonard.popp@student.kit.edu}, \texttt{\{firstname.lastname\}@kit.edu}
        }

\begin{document}
\maketitle
\begin{abstract}
Adapting large language models to an individual author's style from a few examples is challenging, and scientific writing sharpens the difficulty: formal conventions leave little surface variation, and authors write about their own topics, so extracted ``style'' easily entangles with content.
We study style-conditioned abstract generation from a few example abstracts per author and propose three methods: (1) contrastive activation steering, (2) a network that predicts steering vectors, and (3) a hypernetwork that predicts LoRA adapters. 
We find a consistent trade-off between style imitation and output quality: fine-tuning buys most of the available style signal but forfeits fluency, while the hypernetwork achieves the best trade-off on both seen and unseen authors. 
Our steering operates at author level, contrasting an author's abstracts against style-neutral generations for the same content. This holds topic fixed, removes the need for a predefined style inventory, and outperforms inventory-based steering.  
% [EDIT 1a] softened "no single optimal axis" claim
Moreover, our analyses demonstrate that manually extracted and predicted steering vectors are near-orthogonal yet score comparably, indicating that style conditioning here can admit at least two unrelated directions rather than requiring one particular axis.
Our code is available\footnote{\url{https://github.com/MadnessOverflow/scientific-style-gen}}.
\end{abstract}

\section{Introduction}
By in-context learning \citep{DBLP:conf/nips/BrownMRSKDNSSAA20}, large language models can solve a wide range of tasks by conditioning on a few demonstrations without parameter updates. 
Author-level style control, however, is not reliably one of these tasks.
While coarse style attributes such as sentiment or emotion can be controlled through activation-level interventions \citep{subramani-etal-2022-extracting,rimsky-etal-2024-steering,konen-etal-2024-style,zhang-etal-2025-personalized},
imitating an individual author from a handful of texts is harder:
\citet{horvitz-etal-2024-tinystyler} find that an 800M model fine-tuned on authorship embeddings outperforms GPT-4 at few-shot transfer toward specific users.
This is the \textit{cold-start} setting of user-level personalization \citep{zhang2025personalization}:
per-user fine-tuning \citep{tan-etal-2024-democratizing,dan-etal-2025-p} requires a history the new author does not have, 
retrieval over that history is equally unavailable \citep{salemi-etal-2024-lamp,DBLP:journals/corr/abs-2310-20081}, 
and a few examples are often insufficient for the model to extract style \citep{reif-etal-2022-recipe,DBLP:journals/corr/abs-2212-08986}.
Scientific writing exacerbates the difficulty.
Formal conventions leave little surface variation, and authors write in their style about their topics, so a model extracting an author's style risks copying content rather than form \citep{wegmann-etal-2022-author,patel-etal-2025-styledistance}.

Existing approaches address parts of this problem but leave a clear gap.
Activation steering has been evaluated on coarse stylistic distinctions \citep{konen-etal-2024-style} or in settings with substantial per-attribute supervision \citep{rimsky-etal-2024-steering,3737916.3742333}, and \textsc{StyleVector} \citep{zhang-etal-2025-personalized} computes and stores one contrastive vector per user offline.
Adapters generated by hypernetworks \citep{DBLP:journals/corr/HaDL16} remove per-user training, but Text-to-LoRA \citep{charakorn2025texttolora} conditions on natural-language task descriptions, and \textsc{p2p} \citep{tan-etal-2026-instant} on user profiles assembled from history summaries and retrieval.
% Work on low-resource authorship style transfer \citep{krishna-etal-2020-reformulating,DBLP:journals/corr/abs-2212-08986} targets the right granularity but assumes a paraphrase-based pipeline rather than parameter- or activation-level conditioning.
None of these works targets fine-grained, author-level style in a strict few-shot regime.
We study style-conditioned abstract generation from three example abstracts per author, comparing three baselines (zero-shot, few-shot, LoRA) against three proposed methods (\S\ref{sec:methods}): 1) contrastive activation steering, 2) a network predicting steering vectors, and 3) a hypernetwork predicting LoRA adapters. Our contributions are:
\begin{itemize}[nolistsep,leftmargin=*]
    \item We expose a consistent trade-off between style imitation and output quality across all six methods, on which the hypernetwork appears to occupy the most efficient position (\S\ref{subsec:tradeoff}).
% [EDIT 1b] softened contribution claim
\item We show that manually computed and predicted steering vectors are near-orthogonal yet yield comparable style accuracy, so at least two unrelated directions support style conditioning and no particular axis is required (\S\ref{subsec:no_single_axis}).
\item We adapt contrastive steering to the author level by contrasting ground-truth abstracts against style-neutral model generations for the same content, and find it matches or exceeds the method of \citet{konen-etal-2024-style} (\S\ref{subsec:layer_scope}).
\end{itemize}

\section{Related Work}

\paragraph{Few-Shot Personalization of LLMs}
Style-conditioned generation sits at the intersection of text style transfer and user-level personalization \citep{zhang2025personalization}.
\citet{horvitz-etal-2024-tinystyler} target the low-data regime by conditioning a small model on authorship embeddings for few-shot transfer toward individual users.
The strongest personalization results, however, come from per-user fine-tuning that
trains dedicated LoRA adapters \citep{hu2022lora} for each user \citep{tan-etal-2024-democratizing,tan-etal-2024-personalized,dan-etal-2025-p}.
This scales poorly in the number of users and degrades when user histories are limited.
This work targets the cold-start setting at author-level granularity, in scientific writing, 
a domain whose formal constraints leave comparatively little surface variation to exploit.

\paragraph{Activation Steering}
Hidden activations contain latent directions that control stylistic attributes \citep{subramani-etal-2022-extracting,DBLP:conf/aaai/ZhangWLAH25}.
\citet{konen-etal-2024-style} derive style vectors by contrasting mean activations of a target style against a fixed set of remaining styles, at a coarse granularity such as modern versus old English.
We instead contrast an author's ground-truth abstracts against style-neutral abstracts the model generates for the same content, which removes the need for a predefined inventory, holds topic constant, and operates at author level. 
% [EDIT 2b] narrowed novelty framing w.r.t. STYLEVECTOR
The matched-content contrast is close in spirit to \citet{zhang-etal-2025-personalized}; our contribution is its use at author level in scientific writing and, primarily, the prediction of author-specific interventions from three examples.
\citet{zhang-etal-2025-personalized} store one contrastive vector per user, computed offline from that user's texts.
We further train a network to predict steering vectors in a single forward pass from three examples, so no per-user artifact is computed or stored.

\paragraph{Hypernetwork-Generated Adapters}
Hypernetworks \cite{DBLP:journals/corr/HaDL16} generate adaptation parameters directly, removing the need for per-user training: prior work produces attribute-conditioned prefixes for a frozen LM \citep{clive-etal-2022-control} and user-specific LoRA adapters for perspectivist classification \citep{ignatev-etal-2025-hypernetworks}. Text-to-LoRA \citep{charakorn2025texttolora}, whose backbone we adopt, maps a natural-language task description to LoRA weights for zero-shot adaptation to unseen tasks. We instead predict style-conditioned adapters for a fixed task from an aggregated embedding of examples. Closest to our setting, \citet{tan-etal-2026-instant} map an encoded user profile to LoRA weights for long-form generation, but rely on an LLM-generated history summary and a retrieval pipeline, whereas we condition on three abstracts alone.

\section{Task and Evaluation}
\label{sec:task_and_eval}

\begin{figure*}[t]
\centering
\begin{tikzpicture}[
  font=\small,
  box/.style={draw, rounded corners=2pt, inner sep=2pt, align=center},   % <- 3pt -> 2pt
  frozen/.style={box, fill=black!8},
  learn/.style={box, fill=blue!8},
  arr/.style={-{Latex[length=1.8mm]}, semithick},
  grad/.style={-{Latex[length=1.8mm]}, semithick, dashed, blue!60},
]
% ---- band 1: author representation ----
\node (refs) {$R_a$};
\node[frozen, right=4mm of refs] (star) {frozen\\embedder};
\node[learn, right=4mm of star] (proj) {projection\\$+$ mean};
\node[right=4mm of proj] (za) {$\mathbf{z}_a$};
\draw[arr] (refs) -- (star);
\draw[arr] (star) -- (proj);
\draw[arr] (proj) -- (za);
% ---- band 2: backbone ----
\node[learn, right=6mm of za] (mlp) {MLP\\backbone};
\node[learn, below=4mm of mlp] (emb) {$E_{\mathrm{layer}}[l]$\\$E_{\mathrm{module}}[m]$};  % <- 7mm -> 4mm
\draw[arr] (za) -- (mlp);
\draw[arr] (emb) -- (mlp);
% ---- band 3: heads ----
\node[learn, above right=2mm and 8mm of mlp] (h1) {steering\\head};   % <- 3mm -> 2mm
\node[learn, below right=2mm and 8mm of mlp] (h2) {LoRA\\head};       % <- 3mm -> 2mm
\draw[arr] (mlp.east) -- (h1.west);
\draw[arr] (mlp.east) -- (h2.west);
\node[right=4mm of h1] (v)  {$\mathbf{v}^{(l)}$};
\node[right=4mm of h2] (ab) {$\mathbf{A}^{(l)},\mathbf{B}^{(l)}$};
\draw[arr] (h1) -- (v);
\draw[arr] (h2) -- (ab);
% ---- band 4: frozen LM ----
\coordinate (lmanchor) at (ab.east |- mlp);
\node[frozen, right=9mm of lmanchor, minimum height=12mm, minimum width=15mm] (lm)  % <- 15mm -> 12mm
  {frozen\\LLM layer\\[-1pt]\footnotesize $l \in L$};
\draw[arr] (v.east)  -- ([yshift=3mm]lm.west);    % <- 4mm -> 3mm
\draw[arr] (ab.east) -- ([yshift=-3mm]lm.west);   % <- 4mm -> 3mm
% inference output, explicitly labelled
\node[right=10mm of lm] (out) {$\hat{y}$};
\draw[arr] (lm) -- node[above, font=\scriptsize] {inference} (out);
% training loss, explicitly labelled
\node[box, above=8mm of lm, align=center] (loss)                      % <- 11mm -> 8mm
  {$\mathcal{L}_{\mathrm{CE}}$ on abstract $y$\\[-3pt]\scriptsize (teacher forced, prompt masked)};  % <- -1pt -> -3pt
\draw[arr] (lm) -- node[right, font=\scriptsize] {training} (loss);
% gradient: back through the frozen LM, the heads, the backbone, into the projection
\draw[grad] (loss.west) -| node[pos=0.12, above, font=\scriptsize, text=blue!60]
  {gradient} (proj.north);
\draw[grad] (loss.west) -| node[pos=0.12, above, font=\scriptsize, text=blue!60]
  {} (mlp.north);
\draw[grad] (loss.west) -| node[pos=0.12, above, font=\scriptsize, text=blue!60]
  {} (h1.north);
\begin{scope}[on background layer]
\draw[grad] (loss.west) -| node[pos=0.12, above, font=\scriptsize, text=blue!60]
  {} (h2.north);
\end{scope}

\draw[grad] ([xshift=-3pt]mlp.south) -| node[pos=0.12, above, font=\scriptsize, text=blue!60]
  {} ([xshift=-3pt]emb.north);
% ---- LM inputs ----
\node[below=5mm of lm] (s) {paper outline $s$};   % <- 8mm -> 5mm
\draw[arr] (s) -- (lm);
\node[left=7mm of s, draw, dashed, red!65, inner sep=2pt] (no) {$r_1, r_2, r_3$};
\draw[red!65, semithick] (no.north west) -- (no.south east);
\draw[red!65, semithick] (no.north east) -- (no.south west);
\node[left=2mm of no, font=\scriptsize, text=red!65] {not in context};
\end{tikzpicture}
\caption{Both style control networks differ only in their output head.
The author-specific example abstracts $R_a$ are embedded into $\mathbf{z}_a$, concatenated with a layer embedding
(for the hypernetwork, also a module embedding) and passed through the MLP backbone.
The steering head emits a vector added to the residual stream.
The LoRA head emits adapter weights inserted into the attention projections.
The network is queried once per target layer $l \in L$ (and per module $m$ for LoRA prediction).
}
%\vspace{-10pt}
\label{fig:arch}
\end{figure*}

\paragraph{Task Formulation}
Given a target author $a$, a reference set $R_a = \{r_1, r_2, r_3\}$ of three
abstracts by $a$, 
and a \textit{structured outline} $s$ of an unseen paper, 
the task is to generate an abstract $\hat{y} = M(s, R_a)$ that describes $s$ in the style of $a$.
Outlines are produced from the paper's \LaTeX{} source (\S\ref{subsec:data}); no
model observes the original abstract as input.
The base model $M$ is frozen throughout, and no method requires per-author
training at deployment.
The small size of $R_a$ is a property of the cold-start setting~\citep{zhang2025personalization} rather than a
hyperparameter.
Unlike prior work, we assume neither a rich user profile nor a retrieval corpus~\citep{tan-etal-2026-instant}.
We evaluate on authors seen during training and on held-out authors entirely
(\S\ref{subsec:data}).

\paragraph{Author Style Evaluation}
We measure style by authorship attribution.
Instead of a standard classifier which fixes the label set at training time, 
our unseen-author regime requires attributing text to new authors introduced at inference from three examples. 
Therefore, we adopt $k$-NN retrieval in a learned embedding space, 
which admits new authors without retraining.
Specifically, 
an encoder is first fine-tuned on authorship classification. 
Then we discard its head, and create document embeddings by mean-pooling %: $\mathbf{e}_x = \frac{1}{M}\sum_{i=1}^{M} \mathbf{h}_i$.
With the example abstracts forming the reference database, a query is assigned the majority label among its $k$ nearest neighbors under cosine distance, falling back to the nearest neighbor when no majority exists.
We report accuracy and mean reciprocal rank (MRR), 
the latter being continuous and therefore less sensitive to sampling variance.
Note that style and content are not fully separable in this metric, 
as abstracts generated without author information are attributed well above chance (\S\ref{subsec:metric_validation}), 
a known issue of authorship
representations~\citep{wegmann-etal-2022-author, 10.1016/j.knosys.2024.111867}.
So we only use style accuracy to compare models rather than as an absolute metric.

\paragraph{Topic Preservation Evaluation}
The generated abstracts must not deviate from the main text's topic.
Therefore, we detect off-topic generation with a binary LLM-judge. 
\texttt{GPT-OSS-20B}~\cite{openai2025gptoss120bgptoss20bmodel} receives an abstract and the introduction of a paper and
decides whether both originate from the same document (prompt in Appendix~\ref{app:topic_preservation_prompt}).

\paragraph{Text Quality Evaluation}
Besides topic preservation, we evaluate the intrinsic quality of the generated abstracts.
Perplexity is an established metric for text fluency~\citep{Holtzman2020The} but is
untrustworthy here, as it rewards the predictable, repetitive text that aggressive style conditioning may produce.
Instead, we measure quality by pairwise preference by letting \texttt{GPT-OSS-20B} compare each
generated abstract against the author's real abstract for the same paper (prompt in Appendix~\ref{app:win_rate_prompt}).
Each pair is judged twice in swapped order to cancel position
bias~\citep{ko-etal-2020-look,qin-etal-2024-large}.
We report the preference rate against the human-written abstract.
Since our corpus predates 2022 when LLM-assisted writing emerged (\S\ref{subsec:data}) and the judge tends to favor LLM-generated content~\citep{zheng2023judging}, 
preference rates are high across all models.
However, note that it is the differences between the preference rates, not their absolute level, that carry the information.

\section{Style Control Methods} \label{sec:methods}

\paragraph{Manual Contrastive Steering}
Hidden activations contain directions that control stylistic attributes, and
adding a style vector to selected layers steers generation toward the
corresponding style.
Prior work~\citep{konen-etal-2024-style} obtains the steering vector by contrasting a target style against the mean
activation of all other styles in a fixed inventory.
This does not transfer to our case, 
as such an inventory hardly exists in practice, 
and contrasting one author against others writing about different topics leaves content in the vector.
Instead, we contrast each reference abstract against the model's own ``neutral'' abstract for the same paper, 
generated zero-shot from $s_i$ in the model's default style.
Specifically, for layer $l$, the steering vector is
\begin{equation}
\mathbf{v}_a^{(l)} = \bar{\mathbf{a}}_{\text{auth}}^{(l)} - \bar{\mathbf{a}}_{\text{gen}}^{(l)},
\end{equation}
where both terms are activations mean-aggregated over tokens.
The three per-reference vectors are averaged and injected during generation,
\begin{equation}
\mathbf{a}_{\text{steered}}^{(l)}(x) = \mathbf{a}^{(l)}(x) + \lambda\,\mathbf{v}_a^{(l)},
\end{equation}
with intensity $\lambda$ and target layers $L$ selected on validation (\S\ref{subsec:impl_details}).
% [EDIT 2a] declare extra deployment-time inputs of manual steering + reproducibility details
Our approach outperforms the inventory-based method of~\citet{konen-etal-2024-style}, as shown in \S\ref{subsec:layer_scope}.
Three constraints remain.
First, the contrast requires one neutral generation per reference, which in turn requires the outline of the corresponding reference paper.
Manual steering therefore consumes strictly more than $R_a$ and does not satisfy the deployment setting of \S\ref{sec:task_and_eval};
we report it not as a competing method under that setting, but as a reference point the two learned methods must match using the three abstracts alone.
Second, it requires a $\lambda$ search per setting: we sweep $\lambda \in \langle\text{grid}\rangle$ on validation and select $\lambda = 0.875$ for the test configuration, whereas the layer comparison in \autoref{tab:layer_scope} was run at $\lambda = 1.0$ before this search.
Third, the intervention is applied to the residual stream at the $\langle$output of each target block / post-attention position$\rangle$ during both prefill and autoregressive decoding, and the activations in Equation~1 are mean-pooled over the $\langle$abstract tokens only, with prompt tokens excluded$\rangle$.

\paragraph{Learned Steering}
We tackle both constraints by predicting the intervention directly from $R_a$.
Each reference is embedded with a frozen STAR encoder\footnote{
STAR is trained contrastively for authorship and therefore expected to be more stylistic than semantic. 
It also outperformed \href{https://huggingface.co/Qwen/Qwen3-Embedding-0.6B-GGUF}{\texttt{Qwen/Qwen3-Embedding}} in initial experiments.}~\citep{10.1016/j.knosys.2024.111867},
then projected and averaged into an author representation
\begin{equation}
\mathbf{z}_a = \tfrac{1}{|R_a|}\textstyle\sum_{i} \text{Proj}\big(\text{encoder}(r_i)\big). 
\end{equation}
%In this case $\mathbf{z}_a \in \mathbb{R}^{128}$.
Concatenating $\mathbf{z}_a$ with a learned layer embedding forms the input passed to an MLP backbone with residual blocks following~\citet{charakorn2025texttolora} (\autoref{fig:arch};
details in Appendix~\ref{app:arch}).
A linear head projects to the hidden dimension of $M$ and its output is added to
the residual stream,
$\mathbf{a}_{\text{steered}}^{(l)}(x) = \mathbf{a}^{(l)}(x) + \mathbf{v}_a^{(l)}$.
The head is zero-initialized, so the intervention is a no-op at the start of training.
No explicit $\lambda$ is applied, as the network learns per-layer magnitudes.
Despite this flexibility, a downside is that a single vector per layer shifts the residual stream uniformly for
every token, independently of what is being written. 
We tackle this next.

\paragraph{Hypernetwork for LoRA Weights}
Unlike steering vectors, 
LoRA adapters condition their update on the layer input $x$ rather than adding a constant offset,
so the intervention varies with what the model is writing.
Therefore, we extend the previous setup to predict LoRA weights, inspired by~\citet{charakorn2025texttolora} which predicts LoRA weights given novel task descriptions.
Here, we adopt their architecture, 
with a module embedding for $m \in \{\texttt{q\_proj},
\texttt{v\_proj}\}$ concatenated alongside the layer embedding,
% $\boldsymbol{\phi}_{m,l} = \text{concat}[\mathbf{z}_a, E_{\text{layer}}[l], E_{\text{module}}[m]]$,
and four linear heads emit the flattened adapters $\mathbf{A}^{(l)} \in
\mathbb{R}^{r \times d_{\text{in}}}$ and $\mathbf{B}^{(l)} \in
\mathbb{R}^{d_{\text{out}} \times r}$, where $r$ is the LoRA rank.
This modifies the forward pass as
\begin{equation}
\mathbf{a}^{(l)}(x) = \mathbf{W}_0^{(l)}x + \mathbf{B}^{(l)}\mathbf{A}^{(l)}x .
\end{equation}
The module initialization mirrors standard LoRA, 
where the $\mathbf{A}$-heads are initialized with zero weights
and uniform biases, 
and the $\mathbf{B}$-heads with zero weights and biases.
This leaves the adapters as a no-op at initialization and was essential for training stability in our case.
% TODO (reviewer request): state whether standard LoRA scaling ($\alpha/r$) is applied to the predicted adapters.
We treat the set of adapted layers as a design choice, restricting it to a middle-layer subset as it acts as a regularizer and improves downstream metrics
(\autoref{tab:layer_scope}).

\paragraph{Training}
As shown in \autoref{fig:arch}, 
both learned steering and the hypernetworks are trained end-to-end with the
author's ground-truth abstract.
The base model and the encoder stay frozen, and gradients pass through $M$ into the
prediction network alone.
Crucially, $M$ receives the zero-shot prompt: the references never enter its
context, so style is carried entirely by the predicted parameters.

\section{Experimental Setup}
\subsection{Data Creation} \label{subsec:data}
\paragraph{Source and Filtering} Our dataset is built from an arXiv snapshot.
We retain papers with permissive licenses\footnote{Some licenses require mentioning the authors. We include this in \href{https://github.com/MadnessOverflow/scientific-style-gen}{our repository}.} in the \texttt{cs.AI} and \texttt{cs.LG} categories.
Publication year is capped at 2022,
so that ground-truth abstracts cannot have been written by a modern LLM:
the style targets are \textit{guaranteed human}.
Lacking unique author identifiers,
we treat identical author names as one person
and discard authors publishing in over five arXiv categories
as a disambiguation heuristic.
We assume the first author is a paper's primary writer.
Papers without parsable \LaTeX{} source or an introduction are excluded.

\paragraph{Outline Creation} As defined in \S\ref{sec:task_and_eval}, 
the models receive an outline rather than the full paper.
This ensures that minimal style information is available in the input.
We clean the \LaTeX{} source to plain text and prompt Qwen3-4B-Instruct-2507
to produce bullet points under four headings:
Context, Methodology, Results, and Conclusion.
Additional details and prompts are in Appendix~\ref{app:outline_creation}.

\paragraph{Statistics} After filtering, the dataset contains 1,232 papers from 203 authors,
averaging 6.07 papers each (range 5-14).
One paper per author forms the validation set and one the test set, 
and the remaining 826 form the training set.
Each entry pairs a paper's outline with its ground-truth abstract, its introduction,
and three abstracts sampled randomly from the same author's remaining papers.

\paragraph{Unseen Author Test Set} To evaluate generalization, we create a second test set where none of the authors appeared in training for either the generation models or the author classifier.
In this set, the authors have exactly four papers, which guarantees disjointness from the base dataset.
It contains 1,312 papers from 328 authors:
three per author serve as prompt examples and one as the test sample.

\subsection{Baselines}
We compare our methods to three standard ones, with the corresponding prompts in Appendix \ref{app:abstract_gen}.
\paragraph{Zero-Shot} The base model generates an abstract from the outline alone.
It receives no author information, and therefore provides a lower bound on style
accuracy and a quality anchor.
\paragraph{Few-Shot} The zero-shot prompt is extended with the three reference abstracts, and the model is instructed to extract and apply the author's style.
This measures style transfer without parameter updates.
\paragraph{LoRA} The base model is fine-tuned on the few-shot prompt with adapters on
the query and value projections at rank $r=8$, matching the hypernetwork's
adapter budget so that the two are directly comparable.
It represents the strongest style baseline.
Note that here LoRA is \textit{not per-author}, as our data only contain roughly five abstracts per author.
Instead, a single set of LoRA adapters is optimized across all training authors and conditioned at inference on the three author-specific examples.

\subsection{Implementation Details} \label{subsec:impl_details}
All models use \texttt{Qwen3-4B-Instruct-2507}~\cite{DBLP:journals/corr/abs-2505-09388} as the frozen base model $M$
(36 layers, $d = 2{,}560$).
We decode with sampling ($T=0.7$, repetition
penalty $1.05$, top-$p = 0.8$, top-$k=20$) and generate 1,024 tokens at most.
The $k$NN author classifier is a fine-tuned \texttt{distilbert-base-uncased} with
$k = 3$ and cosine distance.
Both learned steering and the hypernetwork use a four-block MLP backbone with an author
representation of 128 dimensions.

Configuration searches use the validation set, while the test set is reserved for
final evaluation.
Manual steering uses our contrast method at layers 18-20 with $\lambda = 0.875$.
Learned steering encodes the three example abstracts with the STAR encoder \cite{10.1016/j.knosys.2024.111867} and predicts a shift vector for all 36 layers.
The hypernetwork uses the same STAR encoder and predicts rank-8 adapters for the
query and value projections, matching the LoRA baseline's rank.
Further implementation and hyperparameter details are in \autoref{app:impl}.

\section{Quantitative Results}

\subsection{Metric Validation} \label{subsec:metric_validation}

\begin{table}[t]
    \centering
    \small
    \begin{tabular}{lccc}
        \toprule
         & Pred.\ Positive & Pred.\ Negative & Total \\
        \midrule
        GT Positive & 1{,}216 & \phantom{0,}16 & 1{,}232 \\
        GT Negative & \phantom{0,00}2 & 1{,}230 & 1{,}232 \\
        \midrule
        Total & 1{,}218 & 1{,}246 & 2{,}464 \\
        \bottomrule
    \end{tabular}
    \caption{Confusion matrix for the topic preservation judge.
    It attains 99.3\% accuracy, misclassifying 18 of 2{,}464 abstract-introduction pairs.
    Negative pairs are constructed by randomly pairing abstracts with unrelated introductions.}
    \label{tab:topic_judge_confusion}
    %\vspace{-10pt}
\end{table}

Before reporting the main results,
we validate the two automatic metrics:
the topic preservation judge and the author classifier.

\paragraph{Topic Preservation Judge}
We first verify that \texttt{GPT-OSS-20B} can reliably detect whether a generated abstract describes the intended paper.
As shown in \autoref{tab:topic_judge_confusion}, 
on an evaluation set of 2,464 abstract-introduction pairs,
the model achieves 99.3\% accuracy,
with 16 false negatives and 2 false positives.
Therefore, we adopt it as a guard against off-topic generation.

\begin{table}[t]
    \centering
    \small
    \begin{tabular}{lllc}
        \toprule
        & Content & Style & Style Acc. (\%) \\
        \midrule
        & \multicolumn{3}{l}{\emph{Target author's papers}} \\
        $(1)$ & \quad target & human (ground truth) & 53.2 \\
        $(2)$ &\quad target & LoRA (target author)  & 49.8 \\
        $(3)$ &\quad target & none (zero-shot)      & 31.0 \\
        \midrule
        & \multicolumn{3}{l}{\emph{Different author's papers}} \\
        $(4)$ &\quad foreign & LoRA (target author) & \phantom{0}0.3 \\
        $(5)$ &\quad foreign & none (zero-shot)     & \phantom{0}0.0 \\
        \midrule
        $(6)$ &\multicolumn{2}{l}{Random guessing}  & \phantom{0}0.5 \\
        \bottomrule
    \end{tabular}
    \caption{Evaluation of the $k$-NN author classifier. 
    Each row is a set of abstracts labeled with the target author, 
    varying the paper (\emph{Content}) and the source of writing style (\emph{Style}). 
    Removing style at fixed content (row~3) costs 22.2 points, yet the remaining 31.0\% is attainable from content alone. 
    Conversely, styling a different author's paper (rows~4-5) falls to chance.}
    \label{tab:metric_probes}
    %\vspace{-10pt}
\end{table}
\paragraph{Content Entangled with Style}
\autoref{tab:metric_probes} shows results of evaluating the $k$-NN style classifier.
First, the classifier attains 53.2\% accuracy on held-out ground-truth abstracts
against random guess accuracy of 0.5\%.
Two experiments establish what it measures.
Removing style while holding content fixed,
by classifying style-neutral abstracts generated from the same outlines,
drops accuracy to 31.0\%. 
The 22.2-point difference is what the classifier loses when author style is removed under this intervention, and the remaining 31.0 is what content alone supplies; neither figure isolates a purely stylistic or purely content-driven component.
This score stays far above chance,
and as these abstracts preserve the paper's topic,
the residual shows the classifier also draws on content,
an entanglement consistent with known difficulties in authorship attribution
\cite{10.1016/j.knosys.2024.111867, wegmann-etal-2022-author}.
Reversing the manipulation, by varying content while holding style conditioning fixed,
sends every model to or below chance,
equally consistent with a classifier that cannot recognize unfamiliar content
and with generators that cannot carry style across topics.

\paragraph{Interpreting Style Accuracy}
The 53.2\% ground-truth level reflects the classifier's attribution ability
rather than a property of the generative models.
More consequentially,
the style-neutral condition above comes from the same zero-shot model
that serves as our first baseline in \S\ref{subsec:main_results},
so its 31.0\% is not a style score but the level attainable from content alone.
We therefore read style accuracy relative to a content-only \textit{reference point} of 31.0\% and a classifier \textit{ceiling} of 53.2\%.
Neither is a strict bound: the few-shot baseline falls below 31.0\% (\S\ref{subsec:main_results}).

% Requires \usepackage{subcaption} and \usepackage{booktabs}
\begin{table}[ht!]
    \centering
    \small
    \setlength\tabcolsep{2.5pt}
    \begin{subtable}{\linewidth}
        \centering
        \begin{tabular}{lcccc}
            \toprule
            Model & Pref. Rate & Topic Pres. & Style Acc. & MRR \\
             & (\%) & (\%) & (\%) &  \\
            \midrule
            Zero-shot   & \textbf{97.9} & 96.6 & 31.0 & 0.40 \\
            Few-shot           & 91.5 & 97.0 & 28.1 & 0.36 \\
            LoRA          & 38.1 & 95.6 & \textbf{49.8} & \textbf{0.55} \\
            \midrule
            Manual steering            & \underline{94.3} & 97.5 & 34.5 & 0.42 \\
            Learned steering  & 89.6 & 96.1 & 35.0 & 0.43 \\
            Hypernet LoRA & 91.3 & 97.0 & \underline{36.5} & \underline{0.46} \\
            \midrule
            Ground truth        & -- & 97.0 & 53.2 & 0.61 \\
            \bottomrule
        \end{tabular}
        \caption{Seen authors test set ($n=203$).
        Style accuracy is bounded below by 31.0, the level attainable from content alone
        (\S\ref{subsec:metric_validation}), 
        and above by the classifier ceiling of 53.2.}
        \label{tab:main_seen}
    \end{subtable}
    %\vspace{3em}
    \begin{subtable}{\linewidth}
        \centering
        \begin{tabular}{lcccc}
            \toprule
            Model & Pref. Rate & Topic Pres. & Style Acc. & MRR \\
              & (\%) & (\%) & (\%) &  \\
            \midrule
            Zero-shot   & \textbf{97.4} & 99.1 & 22.9 & 0.31 \\
            Few-shot           & 90.2 & 99.7 & 18.3 & 0.26 \\
            LoRA          & 39.0 & 97.0 & \textbf{37.5} & \textbf{0.47} \\
            \midrule
            Manual steering            & 95.1 & 98.5 & 26.2 & 0.37 \\
            Learned steering  & 87.5 & 99.4 & 28.4 & 0.38 \\
            Hypernet LoRA  & \underline{95.5} & 99.4 & \underline{31.4} & \underline{0.40} \\
            \midrule
            Ground truth        & -- & 100.0 & 41.5 & 0.49 \\
            \bottomrule
        \end{tabular}
        \caption{Unseen authors test set ($n=328$).
        Neither the models nor the author classifier were trained on any sample from this set.
        Style accuracy is bounded below by 22.9, 
        and above by the classifier ceiling of 41.5.}
        \label{tab:main_unseen}
    \end{subtable}
    \caption{Style and quality results on two tests sets with (a) seen and (b) unseen authors.}
    \label{tab:main_results}
\end{table}
\subsection{Results on Seen Authors}
\label{subsec:main_results}
\autoref{tab:main_seen} reports test set results for our three approaches
alongside the three baselines and the ground truth,
evaluated on the 203 authors seen during training.
We report preference rate as the primary quality metric and author classifier accuracy as the primary style metric.
All models preserve the paper's topic at comparable levels (95.6-97.5\%),
which is a precondition for comparing style accuracy rather than a discriminating result.

\paragraph{Few-shot baseline fails to convey author style.}
The few-shot baseline scores below the zero-shot baseline on both style metrics
(28.1\% and 0.36 against 31.0\% and 0.40),
despite receiving the same examples the LoRA baseline is trained with.
It thus falls below the content-only reference point,
potentially because instruction-following pulls the output toward a generic instructed register, moving embeddings away from any specific author.
Quality degrades as well (91.5\% preference rate).

\paragraph{Direct fine-tuning maximizes style at the cost of quality.}
The LoRA setup attains the highest style accuracy at 49.8\%,
roughly 84\% of the range above the content-only reference point.
However, its preference rate falls drastically to 38.1\% from 97.9\% for the zero-shot baseline,
so fine-tuning purchases nearly all of the available style signal
by sacrificing what makes the abstracts usable.
Manual inspection reveals fluency issues with frequent repetitions. %, and verbose outputs that exceed common abstract lengths.

\paragraph{Activation steering preserves quality but yields small gains.}
Both steering approaches improve style while leaving quality largely intact.
Manual steering reaches 34.5\% at a preference rate of 94.3\%,
the highest of any model improving on the zero-shot style accuracy,
and the prediction network reaches 35.0\% at 89.6\%.
Predicting vectors therefore matches hand-computed ones on style
without a neutral generation per reference or a search over intensity.
The two differ by a single abstract,
and \S\ref{subsec:no_single_axis} shows they achieve this by steering in nearly orthogonal directions.

\paragraph{Hypernet LoRA balances style and quality.}
The hypernetwork LoRA reaches 36.5\%, roughly a quarter of the range,
while retaining a preference rate of 91.3\% that direct fine-tuning forfeits.
Both amortize across authors rather than fitting each separately,
differing in that LoRA carries style through the context window with a shared adapter,
whereas the hypernetwork carries it through adapter weights predicted per author.
Note the two methods are matched on adapter budget but not on information: LoRA reads the example text directly and can copy surface phrasing, while the hypernetwork conditions only on their embeddings.
% Whether the balance reflects an advantage of predicted adapters
% or a weaker intervention cannot be determined here (\S\ref{subsec:layer_scope}).

\subsection{Generalization to Unseen Authors}
\label{subsec:unseen_authors}

So far, the models were tested on authors present in the training data.
We now assess if style conditioning transfers to authors that neither the models nor the author classifier have seen (\autoref{tab:main_unseen}).

\paragraph{Model rankings are preserved.}
Every value is lower than on the seen authors,
including the ground truth (41.5\% against 53.2\%).
Only the relative ordering is informative, and it is identical,
with few-shot again last and below the zero-shot baseline.
As the seen and unseen sets share no authors,
generation noise is an implausible explanation for the few-shot result.

\paragraph{Hypernet retains position.}
The hypernetwork improves style by 8.5 points over the zero-shot baseline
while giving up 1.9 points of preference rate, ranking second on all three metrics,
and the standard LoRA setup again buys the highest style accuracy at a preference rate of 39.0\%.
The trade-off structure thus reappears unchanged
on authors no component of the system was trained on.
The hypernetwork now recovers close to half of the range
against roughly 79\% for LoRA,
where the seen-author fractions were a quarter and 84\%,
though the narrower unseen range inflates these fractions and they are not directly comparable across sets.

% \paragraph{Learned Steering Overtakes Manual Steering on Style but Not Quality}
% Predicted steering exceeds manual steering by seven abstracts on style
% but attains a lower preference rate (87.5\% against 95.1\%),
% a difference our point estimates do not resolve.

\begin{figure}[t]
\centering
\begin{subfigure}[b]{0.45\textwidth}
  \includegraphics[width=1\linewidth]{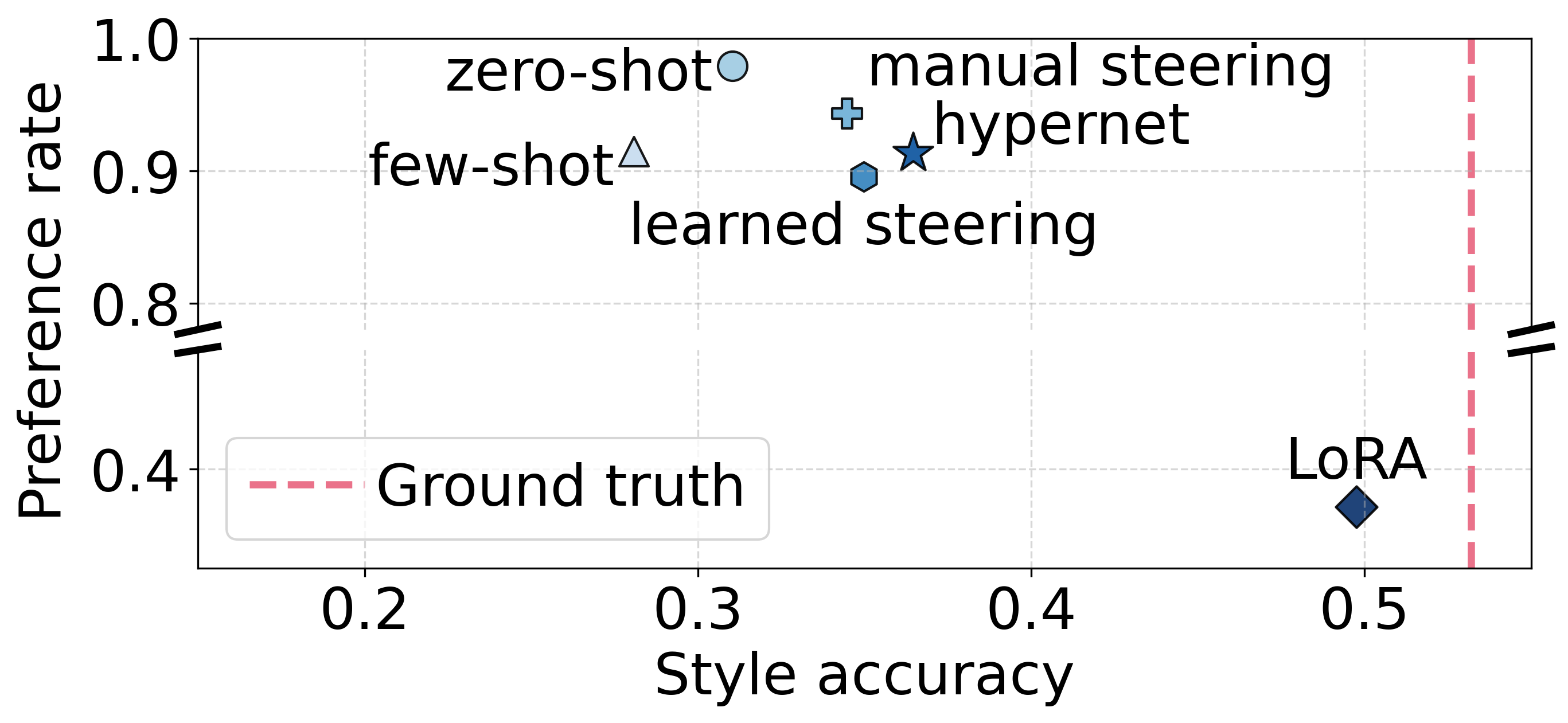}
  \caption{Seen author test set}
  \label{fig:Ng1} 
\end{subfigure}

\medskip
\begin{subfigure}[b]{0.45\textwidth}
  \includegraphics[width=1\linewidth]{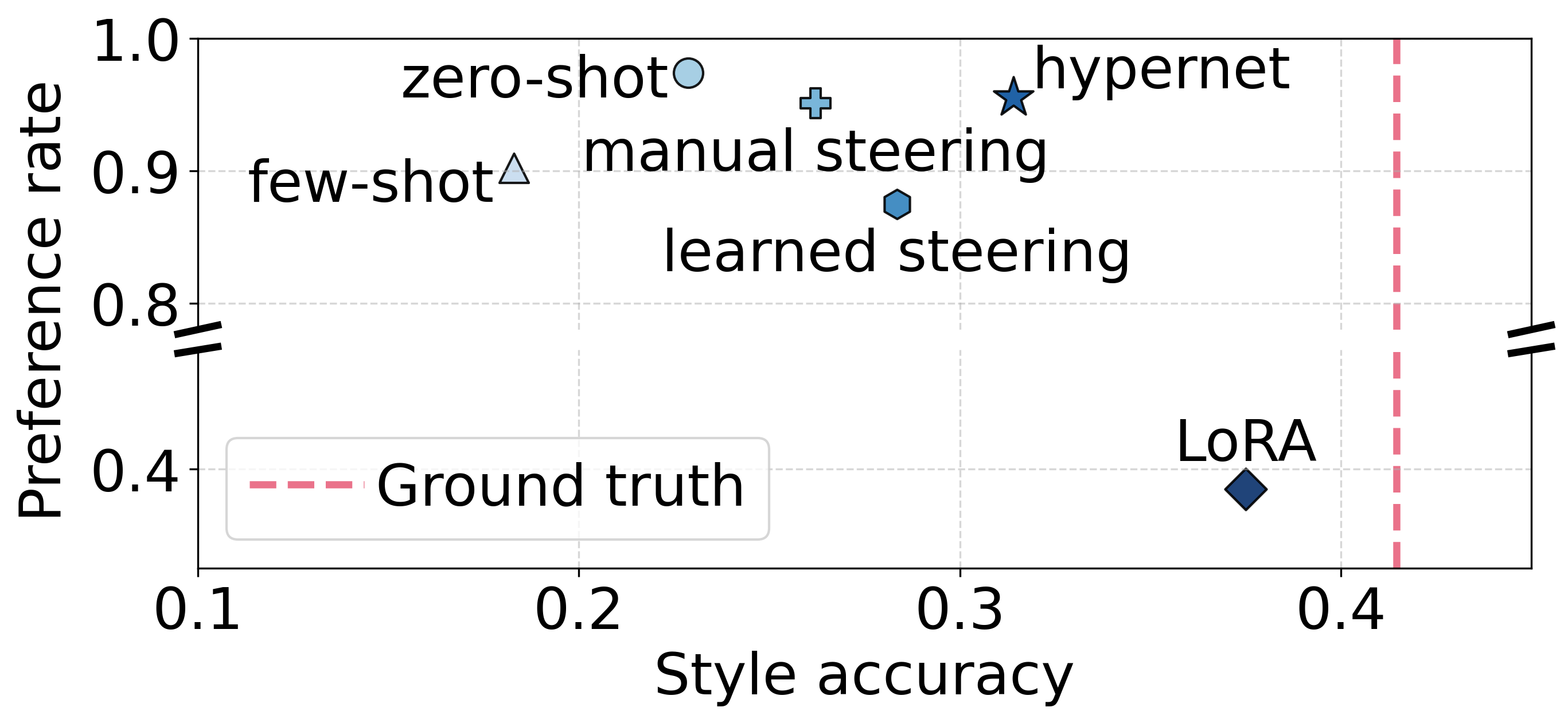}
  \caption{Unseen author test set}
  \label{fig:Ng2}
\end{subfigure}

\caption{\label{fig:tradeoff}%
Style-quality trade-off by plotting the preference rate against style accuracy for each method, on seen (a) and unseen authors (b). 
The hypernetwork sits immediately before quality degrades.}
%\vspace{-10pt}
\end{figure}
\subsection{Style-Quality Trade-Off} \label{subsec:tradeoff}

\autoref{fig:tradeoff} plots preference rate against style accuracy.
% As preference rate is not monotone in style accuracy and the models are sparsely distributed,
% no smooth trade-off curve is identifiable,
% [EDIT 3] common-scope comparison, uncertainty estimates, arithmetic fix (56.6 -> 56.5)
We characterize the trade-off by Pareto dominance on point estimates.
Three models are Pareto-optimal: the zero-shot baseline on preference rate,
LoRA on style accuracy, and the hypernetwork LoRA between them.
On the seen authors the frontier also contains manual steering,
so the hypernetwork is the only one of our approaches Pareto-optimal on \textit{both} sets at the configurations of \autoref{tab:main_seen}.
Two qualifications apply.
First, the two learned methods are reported at different intervention scopes: learned steering over all 36 layers and the hypernetwork over layers 18--20.
At the common scope 18--20 (\autoref{tab:layer_scope}), the hypernetwork retains higher style accuracy (36.5\% vs.\ 34.0\%) but learned steering attains the higher preference rate (97.5\% vs.\ 91.3\%), so neither dominates and the seen-set frontier is scope-dependent.
Second, several style-accuracy differences correspond to few examples; manual and predicted steering differ by a single abstract on the seen-author test set.
We therefore report paired bootstrap 95\% confidence intervals in \autoref{tab:main_seen} and read Pareto statements as descriptive of point estimates rather than as significance claims.
Relative to the hypernetwork, on the unseen set, LoRA buys 6.1 points of style accuracy
for 56.5 points of preference rate.
In contrast, the hypernetwork occupies the frontier between the two, retaining quality while recovering a meaningful share of the style range.

\section{Analyses}
In addition to quantitative results above, we conduct several analyses.

\subsection{No Single Style Axis} \label{subsec:no_single_axis}
\begin{table*}[t]
    \centering
    \small
    \begin{tabular}{lccccccc}
        \toprule
        & \multicolumn{3}{c}{L2 Norm} & \multicolumn{3}{c}{L1 / L2} & Cosine Similarity \\
        \cmidrule(lr){2-4} \cmidrule(lr){5-7} \cmidrule(lr){8-8}
        Layer & Manual & Pred. & Hidden & Manual & Pred. & Hidden & Manual vs.\ Pred. \\
        \midrule
        18 & 13.4 & 14.3 & 39.9 & 17.3 & 40.1 & 11.8 & $-0.03$ \\
        19 & 13.6 & \phantom{0}8.2 & 46.4 & 18.1 & 39.5 & 12.1 & $-0.05$ \\
        20 & 15.1 & 12.2 & 49.4 & 19.4 & 39.7 & 12.0 & $-0.04$ \\
        \bottomrule
    \end{tabular}
    \caption{Per-layer statistics for manual and predicted steering vectors,
    averaged over the 203 validation samples.
    \emph{Hidden} indicates the \textit{unsteered} activations the vectors are added to.
    L1/L2 measures how broadly a vector spreads its mass,
    with 40.4 expected of a dense Gaussian vector in 2,560 dimensions
    and 50.6 of a uniform one.
    The final column reports the average cosine similarity
    between the manual and predicted vector at each layer.}
    \label{tab:vector_comparison}
    %\vspace{-10pt}
\end{table*}
As shown previously in \autoref{tab:main_seen}, manual steering and the activation vector prediction network reach comparable style accuracies.
Since both add a vector to the model's hidden activations,
we ask: do both methods achieve the same \textit{behavior} with the same \textit{internal mechanisms}?
\autoref{tab:vector_comparison} compares the vectors both produce at layers 18 to 20, averaged over 203 validation samples.

\paragraph{Manual and learned steering vectors are near-orthogonal.}
As shown in the last column of \autoref{tab:vector_comparison},
the average cosine similarity between the manual and the predicted vectors
is near zero at every layer ($-0.03$, $-0.05$, $-0.04$).
In 2,560 dimensions this is the value expected of unrelated vectors,
suggesting that the two sets of vectors are therefore \textit{nearly orthogonal}.
% [EDIT 1c] softened claim; grammatical fix (missing verb)
The near-orthogonality combined with their equivalent style accuracy
shows that the predictor does not recover the manually extracted direction,
and that at least two unrelated directions reach comparable style accuracy.
This does not rule out the existence of a privileged direction; it shows that reaching one is not necessary.
Establishing the stronger claim would require comparing vectors extracted from disjoint reference subsets of the same author against vectors from different authors, which we leave to future work.

\paragraph{Manual and learned steering vectors have similar magnitudes but different density.}

Across the two methods, the magnitude of the resulting perturbation is comparable.
As shown in \autoref{tab:vector_comparison}, both produce vectors whose L2 norm falls between 18\% and 36\%
of the corresponding hidden activation norm,
which suggests that a similar quantity of perturbation is required
to influence the model's output regardless of the direction chosen.
Despite comparable magnitudes, 
their \textit{internal structure} does not match.
The ``L1/L2'' column in \autoref{tab:vector_comparison} measures how broadly a vector spreads its mass.
For the predicted vectors it sits at 39.5 to 40.1 at every layer,
matching the value expected of a dense Gaussian vector in 2,560 dimensions.
In contrast, the manual vectors reach only 17.3 to 19.4.
The predicted vectors are therefore \textit{dense},
spreading a small perturbation across many dimensions,
whereas the manual vectors are \textit{sparse},
concentrating the perturbation in a few dominant ones.

\subsection{Layer Selection} \label{subsec:layer_scope}
% \begin{table}[t]
%     \centering
%     \small
%     \setlength\tabcolsep{2.5pt}
%     \begin{tabular}{lccc}
%         \toprule
%         Layers & Style Acc. & MRR & Pref. Rate\\
%         \midrule
%         % \multicolumn{4}{l}{\emph{Manual steering (validation, $\lambda=1.0$)}} \\
%         % \quad 18--20 & 36.0\% & -- & -- \\
%         % \quad all 36 & \phantom{0}2.5\% & -- & -- \\
%         \multicolumn{4}{l}{\emph{Manual steering (validation)}} \\
%         \quad 18--20, ours & 36.0\% & -- & -- \\
%         \quad 20--22, ours & 35.0\% & -- & -- \\
%         \quad 18--20, \citet{konen-etal-2024-style} & 34.0\% & -- & -- \\
%         \quad 20--22, \citet{konen-etal-2024-style} & 31.0\% & -- & -- \\
%         \quad all 36, ours & \phantom{0}2.5\% & -- & -- \\
%         \midrule
%         \multicolumn{4}{l}{\emph{Learned steering (test)}} \\
%         \quad 18--20 & -- & -- & -- \\
%         \quad all 36 & 35.0\% & 0.43 & 89.6\% \\
%         \midrule
%         \multicolumn{4}{l}{\emph{Hypernetwork LoRA (test)}} \\
%         \quad 18--20 & \textbf{36.5\%} & \textbf{0.46} & \textbf{91.3\%} \\
%         \quad all 36 & 34.5\% & 0.43 & 71.4\% \\
%         \bottomrule
%     \end{tabular}
%     \caption{Effect of layer scope on each intervention.
%     Manual steering is reported on the validation set at $\lambda = 1.0$
%     from the configuration search, which measured style accuracy only.}
%     \label{tab:layer_scope}
% \end{table}

\begin{table}[t]
    \centering
    \small
    \setlength\tabcolsep{2.5pt}
    \begin{tabular}{lccc}
        \toprule
        Layers & Style Acc. & MRR & Pref. Rate\\
         & (\%) &  & (\%) \\
        \midrule
        \multicolumn{4}{l}{\emph{Manual steering (validation)}} \\
        \quad 18--20, ours & 36.0 & -- & -- \\
        \quad 20--22, ours & 35.0 & -- & -- \\
        \quad all 36, ours & \phantom{0}2.5 & -- & -- \\
        \quad 18--20, \citet{konen-etal-2024-style} & 34.0 & -- & -- \\
        \quad 20--22, \citet{konen-etal-2024-style} & 31.0 & -- & -- \\
        \midrule
        \multicolumn{4}{l}{\emph{Learned steering (test)}} \\
        \quad 18--20 & 34.0 & 0.44 & \textbf{97.5} \\
        \quad all 36 & \textbf{35.0} & 0.43 & 89.6 \\
        \midrule
        \multicolumn{4}{l}{\emph{Hypernetwork LoRA (test)}} \\
        \quad 18--20 & \textbf{36.5} & \textbf{0.46} & \textbf{91.3} \\
        \quad all 36 & 34.5 & 0.43 & 71.4 \\
        \bottomrule
    \end{tabular}
    \caption{Effect of layer selection.
    Manual steering is reported on the validation set at $\lambda = 1.0$, 
    where the configuration search measured style accuracy only (hence ``$-$'' for MRR and Pref. Rate).
    The learned interventions are reported on the test set.}
    \label{tab:layer_scope}
    %\vspace{-10pt}
\end{table}

\autoref{tab:layer_scope} shows the impact of which layers of the base model our interventions act on.

\paragraph{Manual steering requires a narrow band.}
Holding the contrast method fixed, layers 18-20 give the highest style accuracy
(36.0\%), 
one point above layers 20-22 (35.0\%).
The same ordering holds for the contrast of \citet{konen-etal-2024-style}
(34.0\% vs.\ 31.0\%), which our construction outperforms at both scopes.
Applying the vector to all 36 layers, however, collapses style accuracy to 2.5\%,
far below the level attainable from content alone (\S\ref{subsec:metric_validation}).
A likely explanation is accumulation along the residual stream.
At the layers where we measure it, a single style vector displaces the hidden
state by roughly a third of its norm (\S\ref{subsec:no_single_axis}).
Injecting one at every layer therefore compounds the displacement with depth, and risks driving the state out of the distribution the model was trained on.
That the surviving band is a middle one matches the findings of \citet{konen-etal-2024-style}, 
who locate the same band in a 32-layer model,
and evidence that middle layers carry the most abstract
representations \citep{wendler-etal-2024-llamas,liu-niehues-2025-middle}.
Here, mid-depth likely strikes the best balance, 
as an early perturbation is reshaped by its subsequent layers,
while a late one leaves too few layers to realize it in surface form.

\paragraph{Hypernet prefers a narrow band but survives full depth.}
Compared to predicting adapters for all layers,
restricting to layers 18-20 yields higher style accuracy
(36.5\% vs.\ 34.5\%), MRR (0.46 vs.\ 0.43), and preference rate (91.3\% vs.\ 71.4\%).
We evaluated only these two scopes, 
so the comparison shows that a narrow band beats full depth,
not that 18-20 is optimal for the hypernetwork.
The failure at full depth is milder than for manual steering.
A plausible reason is learning the intervention end-to-end:
Manual steering scales every layer by one global $\lambda$,
while the hypernetwork predicts each layer's adapter separately, 
leaving it free to intervene weakly where a layer tolerates less perturbation, 
as the steering network measurably does (\S\ref{subsec:no_single_axis}).
That a narrow band wins here suggests this freedom is not sufficient on its own. 
Restricting the scope likely acts as a regularizer.

\paragraph{Localization is a property of the intervention.}
Learned steering makes the point more sharply still.
At all 36 layers it reaches 35.0\% style accuracy at an 89.6\% preference rate,
the same scope at which the manual vectors collapse to 2.5\% on validation.
The scope is identical; the vectors are not.
Its behavior across scope also differs from the hypernetwork's:
style accuracy is essentially flat (34.0\% at layers 18--20 vs.\ 35.0\% at full depth, within a point),
while the narrow band instead buys quality, raising the preference rate to 97.5\% from 89.6\%.
Which layers matter, and if narrowing them trades style for quality or improves both,
is thus a property of the specific intervention rather than a fixed rule.

\section{Conclusion}
We studied author-level style conditioning in a strict cold-start setting: three abstracts, no history, no retrieval, and a frozen base model.
Our results lead to three practical recommendations.
(1) Style conditioning here is not a problem of finding the best intervention, but of navigating a trade-off each intervention pays.
% [EDIT 1d] softened conclusion claim
(2) The near-orthogonality of manual and learned steering vectors argues against two practices: extraction is not a search for \textit{the} style direction, since at least two unrelated directions work equally well here and refining the contrast set may only tune an arbitrary choice; and vector similarity cannot validate a method when orthogonal vectors work equally well, only downstream metrics can.
(3) In a contrastive steering vector, the negative should carry the style the output must deviate from, here the model's own default on the same content. %, rather than an average over a fixed inventory.

% % [EDIT 5] ethics statement
% \section*{Ethical Considerations}
% Methods that imitate an individual author's style from three public texts can also be used to impersonate that author, for example to attach a researcher's voice to work they did not write or endorse.
% We use only papers under permissive arXiv licenses, published before 2022, and we condition on abstracts that are already public; no private writing is used, and we release no per-author artifact, since both learned methods produce the intervention on the fly rather than storing a vector or adapter per author.
% We nevertheless note that the same pipeline could be pointed at a non-consenting author, and that our style metric is an attribution model, so improvements on it also indicate increased ability to evade authorship-based detection.
% We regard consent of the imitated author, and disclosure when style-conditioned text is published, as preconditions for applied use.
% \newpage

\section*{Limitations}

Our style metric is entangled with content: style-neutral generations are attributed well above chance, so we can only read accuracy relative to an empirical reference point and ceiling, and comparisons are ordinal rather than absolute. The probe that varies content while fixing style conditioning lands at or below chance, which we cannot attribute to the classifier or to the generators.
Readers should therefore treat every style number as a relative ranking within our setup, never as a measure of how well an author is imitated; disentangling the two would require a classifier trained to be content-invariant and a probe that isolates the generator from the metric.

% [EDIT 4] author-specificity control. Keep the first version if the shuffle control is run;
% otherwise keep the bracketed fallback and delete the first version.
% Our main results do not by themselves establish that the learned interventions are author-specific rather than a generic shift toward human-written abstracts.
% To test this, we repeat the seen-author evaluation with shuffled references: the target outline is held fixed while the intervention is conditioned on three abstracts from a different author.
% Style accuracy falls to $\langle X \rangle$\% for learned steering and $\langle Y \rangle$\% for the hypernetwork, against 35.0\% and 36.5\% with the correct references, so the gain over zero-shot is largely attributable to author identity.
% Consistently, the mean cosine similarity between predicted steering vectors is $\langle A \rangle$ within authors and $\langle B \rangle$ across authors, and the same contrast on the effective hypernetwork update $\mathbf{B}^{(l)}\mathbf{A}^{(l)}$ gives $\langle C \rangle$ and $\langle D \rangle$.
% [FALLBACK if the control is not run:]
Our main results do not by themselves establish that the learned interventions are author-specific rather than a generic shift toward human-written abstracts.
We report the absence of this control as a limitation: without it, the improvement over zero-shot cannot be separated from a generic shift toward human-written abstracts, and the within- versus between-author similarity of predicted vectors and of the effective update $\mathbf{B}^{(l)}\mathbf{A}^{(l)}$ remains unmeasured.

The learned steering and hypernetwork was evaluated at two scopes, so we can only establish that a narrow band beats full depth, not that our selected band is optimal.
Our layer choices should be read as a working configuration rather than a tuned one. 
A full sweep over bands and widths per method is left to future work, and the gap between methods may narrow once each is given its own best scope.

Our methods are run on one base model (Qwen3-4B-Instruct), one broad domain (pre-2022 \texttt{cs.AI/cs.LG} arXiv abstracts), and only English.
Whether the trade-off structure is a property of style conditioning or of this configuration remains open. Replication across model scales, families, domains, and other languages is needed before generalizing.

\section*{Acknowledgment}
We thank the reviewers for helpful comments.
Part of this work was funded by the KiKIT (The Pilot
Program for Core-Informatics at the KIT) of the
Helmholtz Association.

\bibliography{custom}

\appendix

\section{Prompt Templates}
\label{app:prompts}

\subsection{Topic Preservation Judge Prompt}
\label{app:topic_preservation_prompt}

The topic preservation prompt (Figure~\ref{fig:topic_preservation_prompt}) instructs \texttt{GPT-OSS-20B} to compare a generated abstract with the original paper introduction and decide whether they describe the same document, returning a single binary label.
\texttt{GPT-OSS-20B} decodes greedily with the default (medium) reasoning effort.
\begin{figure}[t]
    \centering
    \begin{promptbox}
        {\large\bfseries System Prompt}\par
        \begin{lstlisting}[style=promptstyle]
You are a strict data validation assistant. Your task is to compare an 'Introduction' and an 'Abstract' to determine if they describe the same specific subject matter, research topic, or underlying content.

Output Rules:
1. Return 'True' if the two texts describe the same core content or topic.
2. Return 'False' if the topics are unrelated, different, or contradictory.

Constraint: You must answer with EXACTLY one word: 'True' or 'False'. Do not provide explanations.
        \end{lstlisting}
    \end{promptbox}

    \begin{promptbox}
        {\large\bfseries User Prompt}\par
        \begin{lstlisting}[style=promptstyle]
Please evaluate the following content pair:

### Abstract
{abstract}

### Introduction
{introduction}

Based on the content above, what is the determination (True/False)?
        \end{lstlisting}
    \end{promptbox}
    \caption{The prompt template used to query \texttt{GPT-OSS-20B} to classify the topic preservation of an abstract. \texttt{\{abstract\}} is replaced with the summary of the paper. \texttt{\{introduction\}} is replaced with the introduction the abstract is compared with.}
    \label{fig:topic_preservation_prompt}
\end{figure}

\subsection{Preference Rate Judge Prompt}
\label{app:win_rate_prompt}
The quality evaluation prompt (Figure~\ref{fig:winrate_prompt}) presents \texttt{GPT-OSS-20B} with two abstracts written for the same paper and asks which is of higher qualitative standard, based on clarity, scientific rigor, and narrative flow. 
\texttt{GPT-OSS-20B} decodes greedily with the default (medium) reasoning effort.
For topic preservation the prompt is capped at 3{,}500 tokens; when a sample
exceeds this, only the introduction is truncated from the end, which preserves
the abstract in full since abstracts contain at most 242 words.

\begin{figure}[htbp]
    \centering
    \begin{promptbox}
        {\large\bfseries System Prompt}\par
        \begin{lstlisting}[style=promptstyle]
You are an expert peer reviewer for a top-tier scientific journal. Your task is to compare two abstracts (Abstract A and Abstract B) written for the same paper and determine which one is of higher qualitative standard.

CRITERIA FOR THE WINNER:
Choose the abstract that a mainstream scientific audience would find more compelling, professional, and publishable. Focus on:
1. Clarity & Precision: Which text explains the problem, methodology, and results more clearly and with better scientific vocabulary?
2. Impact & Rigor: Which abstract sounds more objective, authoritative, and convincing?
3. Flow: Which abstract has a better narrative arc that makes the reader want to read the full paper?

OUTPUT FORMAT:
Respond strictly in valid JSON format:
{
  "reasoning": "1-2 sentences explaining why the winning abstract is more convincing to a reviewer.",
  "winner": "<A, B>"
}
        \end{lstlisting}
    \end{promptbox}

    \begin{promptbox}
        {\large\bfseries User Prompt}\par
        \begin{lstlisting}[style=promptstyle]
Please evaluate the following two abstracts.

ABSTRACT A:
{abstract_a}

ABSTRACT B:
{abstract_b}
        \end{lstlisting}
    \end{promptbox}
    \caption{The prompt template used to query \texttt{GPT-OSS-20B} for the quality win rate evaluation. \texttt{\{abstract\_a\}} is replaced with the generated abstract and \texttt{\{abstract\_b\}} with the human-written ground-truth abstract. Each pair is evaluated twice with the input order swapped to eliminate positional bias.}
    \label{fig:winrate_prompt}
\end{figure}

\subsection{Outline Creation Prompt}
\label{app:outline_creation}
Papers exceeding the context window are split into chunks of up to 7,750 tokens with a 250-token overlap,
processed separately, and merged in a second pass.
For papers whose plain text exceeds the model's context limit, 
each chunk is first processed individually using the prompt in Figure~\ref{fig:chunk_summarization_prompt}, which instructs the model to produce a compact bulleted outline of that chunk alone. 
These per-chunk outlines are then combined to form a single, structured outline using the prompt in Figure~\ref{fig:summarization_from_chunks_prompt}. 
For shorter papers that fit within the context window, the paper is directly summarized using the prompt in Figure~\ref{fig:summarization_no_chunks_prompt} without chunking. 
In all cases, the model is a 4-bit quantized version of \texttt{Qwen3-4B-Instruct-2507}.

\begin{figure}[htbp]
    \centering
    \begin{promptbox}
        {\large\bfseries System Prompt}\par
        \begin{lstlisting}[style=promptstyle]
You are a precise scientific summarizer.
Your goal: Synthesize the provided text chunk into a high-density, bulleted summary.
Constraints:
1. Max 400 words. Focus on maximum information density.
2. Preserve all key details important for writing an abstract or summary of the paper.
3. Remove redundancies and repetition.
4. Maximize information density: Prioritize factual content over linguistic fluency or reading flow.
5. Use a dry, compact, and technical style. Avoid flowery language or unnecessary transition words.
6. Especially do NOT use introductory filler (e.g. "This section discusses...").
7. Output ONLY the summary text.
        \end{lstlisting}
    \end{promptbox}

    \begin{promptbox}
        {\large\bfseries User Prompt}\par
        \begin{lstlisting}[style=promptstyle]
Generate a comprehensive bulleted summary for this section:

{text_chunk}
        \end{lstlisting}
    \end{promptbox}
    \caption{The prompt template used to query \texttt{Qwen3-4B-Instruct-2507 4-bit} for chunk summarizations. \texttt{\{text\_chunk\}} is replaced with a chunk of the paper.}
    \label{fig:chunk_summarization_prompt}
\end{figure}

\begin{figure}[htbp]
    \centering
    \begin{promptbox}
        {\large\bfseries System Prompt}\par
        \begin{lstlisting}[style=promptstyle]
You are a senior scientific editor. 
Your goal: Synthesize provided informations into a high-density, bulleted summary that contains every detail necessary to write a formal abstract.

Constraints:
1. Length: 400-500 words. Focus on maximum information density.
2. Structure: Use exactly four bulleted sections: CONTEXT (problem/objective), METHODOLOGY (design/sample/tools), RESULTS (specific findings/data), and CONCLUSION (significance/implications).
3. Requirements: DO NOT omit specific metrics, sample sizes, or p-values. Ensure the logical thread from hypothesis to implication is intact.
4. Use bullet points only.
5. Format: PLAIN TEXT ONLY. No Markdown headers (##). No intro/outro filler. Bold the section labels.
        \end{lstlisting}
    \end{promptbox}

    \begin{promptbox}
        {\large\bfseries User Prompt}\par
        \begin{lstlisting}[style=promptstyle]
The following are summaries of a scientific paper's individual sections:

Section {1} Summary:
{summary_1}

Section {2} Summary:
{summary_2}

[...]

Section {x} Summary:
{summary_x}

---

Generate a comprehensive bulleted summary for these sections now. Ensure that no critical methodological detail or key finding is omitted, as this summary will serve as the sole basis for a formal abstract.
        \end{lstlisting}
    \end{promptbox}
    \caption{The prompt template used to query \texttt{Qwen3-4B-Instruct-2507 4-bit} for papers that exceed the context window and are first split into chunks. \texttt{\{summary\_x\}} is replaced with the summary of chunk $x$. The \texttt{[...]} notation is a schematic abbreviation; the actual prompt contains one numbered section block per chunk.}
    \label{fig:summarization_from_chunks_prompt}
\end{figure}

\begin{figure}[htbp]
    \centering
    \begin{promptbox}
        {\large\bfseries System Prompt}\par
        \begin{lstlisting}[style=promptstyle]
You are a senior scientific editor. 
Your goal: Synthesize provided informations into a high-density, bulleted summary that contains every detail necessary to write a formal abstract.

Constraints:
1. Length: 400-500 words. Focus on maximum information density.
2. Structure: Use exactly four bulleted sections: CONTEXT (problem/objective), METHODOLOGY (design/sample/tools), RESULTS (specific findings/data), and CONCLUSION (significance/implications).
3. Requirements: DO NOT omit specific metrics, sample sizes, or p-values. Ensure the logical thread from hypothesis to implication is intact.
4. Use bullet points only.
5. Format: PLAIN TEXT ONLY. No Markdown headers (##). No intro/outro filler. Bold the section labels.
        \end{lstlisting}
    \end{promptbox}

    \begin{promptbox}
        {\large\bfseries User Prompt}\par
        \begin{lstlisting}[style=promptstyle]
The following is the paper you are required to summarize:

PAPER START:

{content}

---

Generate a comprehensive bulleted summary for these sections now. Ensure that no critical methodological detail or key finding is omitted, as this summary will serve as the sole basis for a formal abstract.
        \end{lstlisting}
    \end{promptbox}
    \caption{The prompt template used to query \texttt{Qwen3-4B-Instruct-2507 4-bit} for papers that fit within the model's context window. \texttt{\{content\}} is replaced with the plain text of the paper.}
    \label{fig:summarization_no_chunks_prompt}
\end{figure}

\subsection{Abstract Generation Prompt}
\label{app:abstract_gen}

Two prompt templates are used for abstract generation, depending on whether the model receives author style examples. The zero-shot template (Figure~\ref{fig:non_icl_prompt}) is used for the zero-shot baseline, the manual activation steering model, the steering vector prediction network and the hypernetwork model. It instructs the model to write a formal academic abstract from a structured paper outline, without any information about the target author. The few-shot template (Figure~\ref{fig:icl_prompt}) is used for the few-shot baseline and the LoRA model. 
It extends the zero-shot template by providing three example abstracts by the target author, instructing the model to adopt that author's style while remaining faithful to the outline content.

\begin{figure}[htbp]
    \centering
    \begin{promptbox}
        {\large\bfseries System Prompt}\par
        \begin{lstlisting}[style=promptstyle]
You are an expert academic writer and stylist. 
Your task is to draft a formal academic abstract by synthesizing information from a summary while strictly adopting a clear, professional, and objective academic tone.

Guidelines:
1. Composition: Construct a cohesive abstract based on the provided 'Summary'. 
2. Style: Apply standard academic vocabulary, precise sentence structures, and logical rhetorical patterns typical of high-quality research papers.
3. Fact Integrity: Use only the information provided in the summary. Do not hallucinate data, results, or citations not present in the source text.
4. Professionalism: Ensure the output follows standard academic conventions for abstracts (objective, concise, and impactful).
5. Constraints: Output only the final abstract. Do not include anything else like headers, remarks, meta-comments, or explanations.
        \end{lstlisting}
    \end{promptbox}

    \begin{promptbox}
        {\large\bfseries User Prompt}\par
        \begin{lstlisting}[style=promptstyle]
Please compose a formal academic abstract based on the following summary.

<summary>
{summary}
</summary>

Output only the final abstract without a header:
        \end{lstlisting}
    \end{promptbox}
    \caption{The prompt template used for zero-shot, manual steering, learned steering, and hypernetwork setup. \texttt{\{summary\}} is replaced with the oueline of the paper.}
    \label{fig:non_icl_prompt}
\end{figure}

\begin{figure}[htbp]
    \centering
    \begin{promptbox}
        {\large\bfseries System Prompt}\par
        \begin{lstlisting}[style=promptstyle]
You are an expert academic writer and stylist. 
Your task is to draft a formal academic abstract by synthesizing information from a summary while strictly adopting the linguistic style, tone, and structural preferences of example abstracts from a specific Author.

Guidelines:
1. Composition: Construct a cohesive abstract based on the provided 'Summary'. 
2. Analyze the 'Author Examples' for sentence structure, vocabulary complexity, tone, and flow.
3. Style Mimicry: Apply the specific vocabulary, sentence structure, and rhetorical patterns associated with the given author examples.
4. Fact Integrity: Use only the information provided in the summary. Do not hallucinate data, results, or citations not present in the source text.
5. Professionalism: Ensure the output follows standard academic conventions for abstracts (objective, concise, and impactful).
6. Constraints: Output only the final abstract. Do not include introductory remarks, meta-comments, or explanations.
        \end{lstlisting}
    \end{promptbox}

    \begin{promptbox}
        {\large\bfseries User Prompt}\par
        \begin{lstlisting}[style=promptstyle]
Here are the examples of the author's writing style:

<style_examples>
Example 1:
{example_1}

Example 2:
{example_2}

Example 3:
{example_3}
</style_examples>

Based on the style of the examples above, compose a formal academic abstract based on the following summary, written in the characteristic style of the specified examples.

<summary>
{summary}
</summary>

Output only the rewritten abstract:
        \end{lstlisting}
    \end{promptbox}
    \caption{The prompt template used for the few-shot and LoRA. \texttt{\{summary\}} is replaced with the outline of the paper. \texttt{\{example\_1\}}, \texttt{\{example\_2\}} and \texttt{\{example\_3\}} are replaced with the example abstracts of the author.}
    \label{fig:icl_prompt}
\end{figure}

\section{Implementation Details}
\label{app:impl}

\paragraph{Environment}
All models are loaded with Unsloth and
Transformers~\citep{wolf-etal-2020-transformers}, and $k$-NN retrieval uses
\textsc{Faiss}~\citep{11202651}.
Every experiment runs on a single NVIDIA TITAN RTX (24\,GB VRAM) with an Intel
Xeon Silver 4110 CPU and 376\,GiB of RAM.
Qwen3-4B-Instruct-2507 is loaded in full precision for generation; the judge is
loaded in 4-bit.
The longest prompt in the dataset is 1{,}851 tokens for outline-only inputs and
2{,}740 tokens with three in-context examples, both well within the context
limit, so no input truncation occurs.

\paragraph{Author Classifier}
\label{app:classifier}
The task is multiclass classification from an abstract to its first author.
We search over four encoders (\texttt{bert-base-uncased},
\texttt{bert-base-multilingual-cased}, \texttt{distilbert-base-uncased},
\texttt{distilbert-base-multilingual-cased}) and three modes: a classification
head, $k$-NN on frozen embeddings, and $k$-NN on fine-tuned embeddings with the
head removed.
For the head mode we search six learning rates
($1{\times}10^{-6}$ to $5{\times}10^{-5}$), selecting by validation loss.
For the retrieval modes we search two distance functions (L2, cosine) and
$k \in \{1,3,5,7,9\}$, selecting by validation accuracy.
Training runs for up to 100 epochs with batch size 16 and AdamW, evaluating each
epoch, with early stopping after five epochs without improvement and restoration
of the best checkpoint.
All abstracts fit within the 512-token limit, so no truncation is applied.
The selected configuration is \texttt{distilbert-base-uncased}, fine-tuned
$k$-NN, $k = 3$, cosine distance, learning rate $2{\times}10^{-5}$.
The $k$-NN database holds all training abstracts; for test-set evaluation the
validation abstracts are added, and for the unseen-author set it holds the three
reference abstracts per author.

\paragraph{LoRA Baseline}
Adapters are applied to \texttt{q\_proj} and \texttt{v\_proj} with $r = 8$ and
$\alpha = 16$.
We use AdamW with a learning rate of $2{\times}10^{-4}$, 5 warmup steps, weight
decay $0.001$, and NEFTune~\citep{jain2024neftune} with noise $\alpha = 5.0$.
The effective batch size is 8 (per-device 2, 4 gradient accumulation steps), and
sequences are padded or truncated to 4{,}096 tokens.
Cross-entropy is computed on the abstract only, with the prompt masked.
Training runs for at most 50 epochs, evaluating every 25 steps, with early
stopping after 25 evaluations without validation-loss improvement and
restoration of the best checkpoint.

\paragraph{Style Control Networks}
\label{app:arch}
The author representation is produced by a frozen STAR encoder; a linear
projection maps its 1{,}024-dimensional output to 128 dimensions before
averaging over the three references.
The layer encoder maps the layer index to a 36-dimensional embedding followed by
layer normalization; the module encoder (hypernetwork only) maps
$\{\texttt{q\_proj}, \texttt{v\_proj}\}$ to a 32-dimensional embedding.
Input dimensions are therefore $128 + 36 = 164$ for the steering network and
$128 + 36 + 32 = 196$ for the hypernetwork.
The backbone is a stack of four MLP blocks with a 512-dimensional
up-projection, SiLU activations, and 5\% dropout: the first block omits layer
normalization, the second and third are residual blocks with pre-normalization,
and the fourth normalizes without a residual connection.
Backbone weights use Kaiming normal initialization with zero biases.
The steering network has a single head projecting $512 \rightarrow 2{,}560$,
zero-initialized in weights and bias.
The hypernetwork has four heads emitting flattened adapters at $r = 8$, with
shapes following Equation~4, i.e.\ $\mathbf{A} \in \mathbb{R}^{r \times d_{\text{in}}}$
and $\mathbf{B} \in \mathbb{R}^{d_{\text{out}} \times r}$:
$\texttt{q}_A$ ($8 {\times} 2{,}560$), $\texttt{q}_B$ ($4{,}096 {\times} 8$),
$\texttt{v}_A$ ($8 {\times} 2{,}560$), and $\texttt{v}_B$
($1{,}024 {\times} 8$).
The $A$-heads have zero weights and biases drawn uniformly from
$[-1/\sqrt{d_{\text{in}}}, 1/\sqrt{d_{\text{in}}}]$; the $B$-heads are zero in
both, matching standard LoRA initialization.
% TODO (reviewer request): state whether standard LoRA scaling ($\alpha/r$) is applied to the predicted adapters.
Training follows the LoRA baseline's configuration, with mixed precision (fp16)
and gradient checkpointing added to reduce memory.

Listings~\ref{lst:backboneArchitectureInput}--\ref{lst:backboneArchitectureOutput} show the full architecture of the shared backbone used by both the activation vector prediction network and the hypernetwork model. 
The input module (Listing~\ref{lst:backboneArchitectureInput}) encodes the author style embedding, the target layer index, and, for the hypernetwork, the target module. The backbone (Listing~\ref{lst:backboneArchitecture}) consists of four sequential MLP blocks. The output heads (Listing~\ref{lst:backboneArchitectureOutput}) differ between the two models: the hypernetwork uses the four LoRA matrix heads, while the activation vector prediction network uses only the single steering vector head.

\newpage
\noindent\begin{minipage}{\linewidth}
\begin{lstlisting}[style=codestyle, language=Python, caption={Input to the style control networks. For the activation vector prediction network, the module embedding is removed. All inputs are concatenated to form the final input to the network. \texttt{num\_of\_layers} is replaced with the number of layers that the model is controlling.}, label={lst:backboneArchitectureInput}]
Input(
  (star_proj): Sequential(
    (0): Linear(in_features=1024, out_features=128, bias=True)
    (1): LayerNorm((128,), eps=1e-05, elementwise_affine=True)
  )
  (layer_depth_encoder): Sequential(
    (0): Embedding({num_of_layers}, 36)
    (1): LayerNorm((36,), eps=1e-05, elementwise_affine=True)
  )
  (module_emb): Sequential(
    (0): Embedding(2, 32)
    (1): LayerNorm((32,), eps=1e-05, elementwise_affine=True)
  )
)
\end{lstlisting}
\end{minipage}

\noindent\begin{minipage}{\linewidth}
\begin{lstlisting}[style=codestyle, language=Python, caption={Output of the style control networks. Either the first four output heads are used for the LoRA hypernetwork or the \texttt{final\_steering} output head is used for the activation vector prediction network. Each head emits a flattened matrix; the reshaped targets follow Equation~4.}, label={lst:backboneArchitectureOutput}]
Output(
  (final_proj_q_A): Linear(in_features=512, out_features=20480, bias=True)
  (final_proj_q_B): Linear(in_features=512, out_features=32768, bias=True)
  (final_proj_v_A): Linear(in_features=512, out_features=20480, bias=True)
  (final_proj_v_B): Linear(in_features=512, out_features=8192, bias=True)
  
  (final_steering): Linear(in_features=512, out_features=2560, bias=True)
)
\end{lstlisting}
\end{minipage}

\newpage
\noindent\begin{minipage}{\linewidth}
\begin{lstlisting}[style=codestyle, language=Python, caption={Backbone of the style control networks. The input is passed through the network in the order of the listing. \texttt{hidden\_dim} is $196$ for the LoRA hypernetwork and $164$ for the activation prediction network.}, label={lst:backboneArchitecture}]
ModuleDict(
  (mlp1): MLPResidualBlock(
      (mlp): Sequential(
        (0): Linear(in_features={hidden_dim}, out_features=512, bias=True)
        (1): SiLU()
        (2): Dropout(p=0.05, inplace=False)
        (3): Linear(in_features=512, out_features={hidden_dim}, bias=True)
        (4): SiLU()
        (5): Dropout(p=0.05, inplace=False)
      )
  )
  (mlp2): MLPResidualBlock(
    (mlp): Sequential(
      (0): LayerNorm(({hidden_dim},), eps=1e-05, elementwise_affine=True)
      (1): Linear(in_features={hidden_dim}, out_features=512, bias=True)
      (2): SiLU()
      (3): Dropout(p=0.05, inplace=False)
      (4): Linear(in_features=512, out_features={hidden_dim}, bias=True)
      (5): SiLU()
      (6): Dropout(p=0.05, inplace=False)
    )
  )
  (mlp3): MLPResidualBlock(
    (mlp): Sequential(
      (0): LayerNorm(({hidden_dim},), eps=1e-05, elementwise_affine=True)
      (1): Linear(in_features={hidden_dim}, out_features=512, bias=True)
      (2): SiLU()
      (3): Dropout(p=0.05, inplace=False)
      (4): Linear(in_features=512, out_features={hidden_dim}, bias=True)
      (5): SiLU()
      (6): Dropout(p=0.05, inplace=False)
    )
  )
  (mlp4): Sequential(
    (0): LayerNorm(({hidden_dim},), eps=1e-05, elementwise_affine=True)
    (1): Linear(in_features={hidden_dim}, out_features=512, bias=True)
    (2): SiLU()
    (3): Dropout(p=0.05, inplace=False)
    (4): Linear(in_features=512, out_features=512, bias=True)
    (5): SiLU()
  )
)
\end{lstlisting}
\end{minipage}

\end{document}